\documentclass[conference]{IEEEtran}
\IEEEoverridecommandlockouts
\usepackage{cite}
\usepackage{amsmath,amssymb,amsfonts}
\usepackage{algorithmic}
\usepackage{graphicx}
\usepackage{textcomp}
\usepackage{xcolor}
\def\BibTeX{{\rm B\kern-.05em{\sc i\kern-.025em b}\kern-.08em
    T\kern-.1667em\lower.7ex\hbox{E}\kern-.125emX}}
\begin{document}

\title{Terrain-adaptive Central Pattern Generators with Reinforcement Learning for Hexapod Locomotion\\
\thanks{*Corresponding author.}
}

\author{\IEEEauthorblockN{1\textsuperscript{st} Qiyue Yang}
\IEEEauthorblockA{\textit{Department of Automation} \\
\textit{Shanghai Jiao Tong University}\\
Shanghai, China \\
yangqiyue@sjtu.edu.cn}
\and
\IEEEauthorblockN{2\textsuperscript{th} Yue Gao}
\IEEEauthorblockA{\textit{MoE Key Lab of Artificial Intelligence and AI Institute} \\
\textit{Shanghai Jiao Tong University}\\
Shanghai, China \\
yuegao@sjtu.edu.cn}
\and
\IEEEauthorblockN{3\textsuperscript{th} Shaoyuan Li*}
\IEEEauthorblockA{\textit{Department of Automation} \\
\textit{Shanghai Jiao Tong University}\\
Shanghai, China \\
syli@sjtu.edu.cn}
}

\maketitle

\begin{abstract}

Inspired by biological motion generation, central pattern generators (CPGs) is frequently employed in legged robot locomotion control to produce natural gait pattern with low-dimensional control signals. However, the limited adaptability and stability over complex terrains hinder its application. To address this issue, this paper proposes a terrain-adaptive locomotion control method that incorporates deep reinforcement learning (DRL) framework into CPG, where the CPG model is responsible for the generation of synchronized signals, providing basic locomotion gait, while DRL is integrated to enhance the adaptability of robot towards uneven terrains by adjusting the parameters of CPG mapping functions. The experiments conducted on the hexapod robot in Isaac Gym simulation environment demonstrated the superiority of the proposed method in terrain-adaptability, convergence rate and reward design complexity. 
\end{abstract}


\section{Introduction}
\label{introduction}
In recent decades, significant advancements have been made in the field of legged robotics \cite{gao2023multi}, which have drawn inspirations from locomotion of legged animals. Examples of such robots include biped robots\cite{li2021reinforcement}, quadruped robots\cite{yang2020data}, and hexapod robots\cite{tian2022capplanner}. With discrete footholds and redundant degrees of freedom, legged robots exhibit remarkable agility in adjusting their body postures, making them well-suited for traversing uneven terrains. Mainstream research methods employed in legged robot locomotion include bio-inspired methods\cite{chen2017control}, data-driven methods\cite{yang2020multi}, and model-based motion optimization methods\cite{liu2021trajectory}.

Central pattern generator, in the first place, refers to the biological neural circuits that produce stable rhythmic actions in the absence of external sensory feedback. It is the underlying mechanism responsible for many spontaneous and unconscious rhythmic behaviors in vertebrates and legged invertebrates, such as swimming, breathing, and walking. The application of CPG in robot control was initially introduced in the motion pattern generation for fish-like and amphibian robot\cite{ijspeert2007swimming}, owing to their natural sinusoidal movements. Subsequently, CPG is applied in legged robot locomotion control, being able to generate natural and bionic gaits with low-dimensional control signals. However, the poor adaptability towards environment limits the further application of this bio-inspired method. To deal with this deficiency, many researchers resort to the combination of structural design \cite{hyun2014high} and sensory feedback \cite{righetti2008pattern} to produce dynamic locomotion on uneven terrains. For instance, in order to realize adaptive walking control on sloped terrains, a three-layered CPG is presented in \cite{liu2018multi} for rhythmic signals generation, motion modes determination and trajectory generation of controlled objects. Wang et al. employs the pitch angle of robot body as feedback to adjust the output characteristics of the signal generator model \cite{wang2020gait}. 

Another promising avenue of research in legged robotics is the data-driven method represented by deep reinforcement learning (DRL). DRL has exhibited remarkable adaption abilities towards various types of terrains \cite{lee2020learning,gangapurwala2022rloc} and external disturbances \cite{yang2020multi}. However, directly learning the mapping from exteroceptive and proprioceptive measurements to joint positions is challenging, which demands careful parameter tuning, skillful reward design and massive data collection \cite{rudin2022learning}. The high-dimensional action space makes the training process difficult to converge, especially for robots with high degrees of freedom like hexapod robots. Additionally, imitation learning methods such as behavior cloning are needed for the acquisition of more biomimetic motion patterns \cite{bong2022standing} . 

Recent works are exploring the potential to strike a balance between the simplicity of bio-inspired methods and the adaptability of data-driven methods, proposing CPG-RL approaches \cite{ouyang2021adaptive,bellegarda2022cpg,li2023combined}. For example, \cite{ouyang2021adaptive} designs a 3D two-layer CPG network consisting of 18 Hopf oscillators, with each oscillator corresponding to a joint actuator of the hexapod robot. The policy network outputs are then used to directly modulate the signal amplitude and phase difference parameter of limb layer. In their work, the outputs of Hopf oscillators are mapped to joint positions, while other studies ~\cite{bellegarda2022cpg,li2023combined} elect to compute corresponding desired foot positions and then transform these positions into desired joint positions though inverse kinematics. These methods have been proved to be successful in generating locomotion on soft sand \cite{ouyang2021adaptive} and rough ground\cite{li2023combined}. However, the terrain-adaptability of these works mostly comes from the natural generalization ability of reinforcement learning and CPG, so they can only complete tests on relatively simple and regular terrains such as flat grounds with different surface friction and minor obstacles.

\begin{figure}
\centerline{\includegraphics[width=0.39\textwidth]{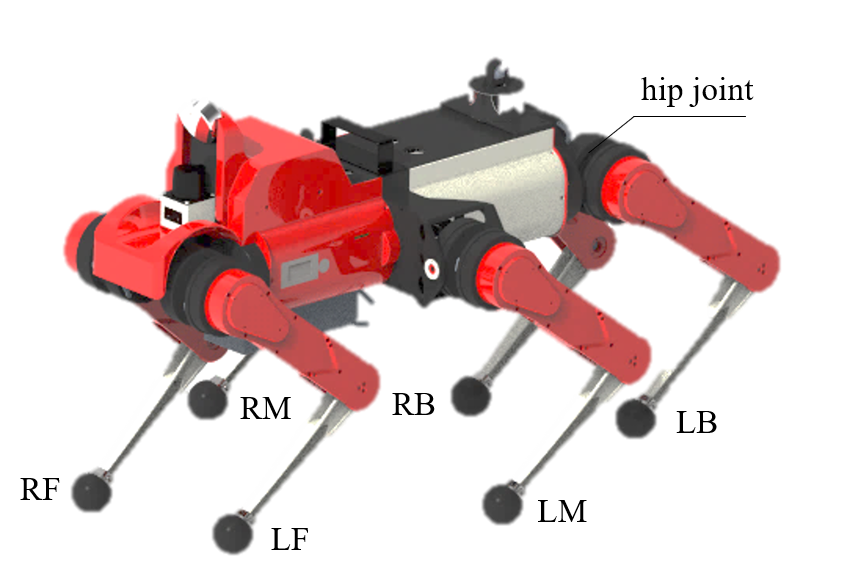}}
\caption{Illustration of the hexapod robot. }
\label{littledog}
\end{figure}

In this work, we are determined to improve the adaptability of hexapod robot towards various terrains. The proposed locomotion control method combines DRL and CPG to leverage their strengths and compensate for their weaknesses. The coupling of oscillating signals and symmetric structure of the robot reduces the dimension of action space, making the training process easier to converge. The application of DRL adds to the adaptability of the robot to different terrains. Specifically, the proposed method involves six intercoupled Hopf oscillators that generate rhythmic signals, which are subsequently mapped to foot positions in the hip coordinate through a mapping function, whose parameters are determined by reinforcement learning. Finally, these foot positions will be transformed into desired joint positions through inverse kinematics module. The contributions of this work include:
\begin{itemize}
    \item a locomotion control approach is proposed, integrating DRL into CPG framework through adjusting the mapping function between state variables of CPG and footholds.
    \item the proposed method is demonstrated to outperform CPG in terrain-adaptability, being able to handle random terrain with height difference of $0.06$ m, sloped terrain with slope of $15^{\circ}$ and wave terrain with amplitude of $0.5$ m.
    \item the dimension of action space and the dependence of the algorithm on reward design is significantly reduced compared with common RL settings, making the training process simpler and easier to converge.
\end{itemize}
The rest of this paper is organized as follows: Section \ref{Background} provides a brief introduction to the robot platform and essential background knowledge about CPG and DRL. Section \ref{details} explains the details on how to integrate DRL into CPG framework through the design of mapping function and Markov Decision Process, and experimental results are presented in Section \ref{Experiments}. Finally, Section \ref{Conclusion} makes a summary of current work and outlines future plans.

\section{Background}
\label{Background}
\subsection{Hexapod Robot}
The robot used in this paper is a hexapod robot. As illustrated in Fig.\ref{littledog}, it comprises six legs, namely right front (RF), right middle (RM), right back (RB), left back (LB), left middle (LM), and left front (LF), each of which possesses 3 degrees of freedom (18 DoFs in total).
\subsection{Central Pattern Generator}
\begin{figure}
\centerline{\includegraphics[width=0.35\textwidth]{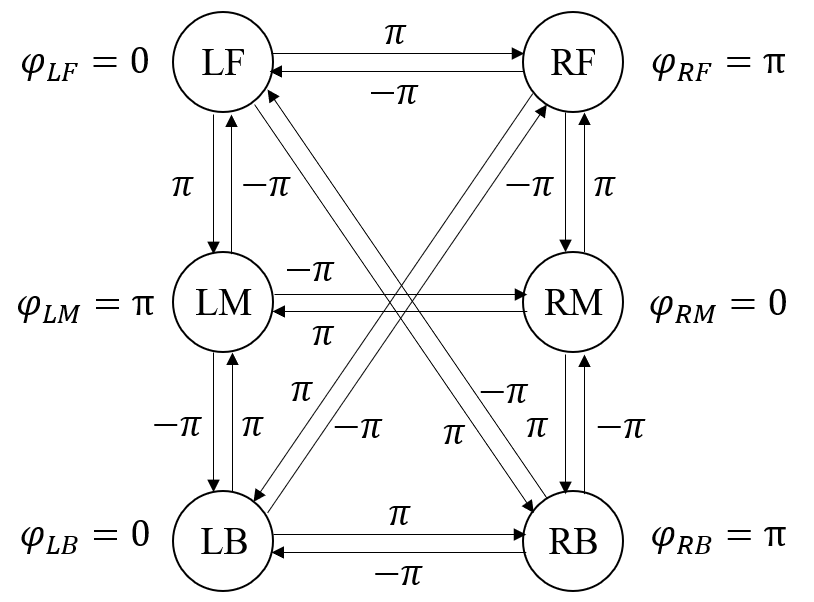}}
\caption{Phase differences of tripod locomotion.}
\label{leg_phase}
\end{figure}

At present, typical CPG models used in robot motion control include neuron-based models and nonlinear oscillator-based models. The former class of models encompasses well-known models such as Matsuoka neural oscillator \cite{endo2008learning} and Kimura model \cite{kimura2007adaptive}. Although they have clearer biological meanings, the non-linearity and high-dimensionality of equations make parameter tuning a challenging task. In contrast, the second category of CPG models, which is based on nonlinear oscillators, is well-known for its simple formulation, few parameters and high stability. Prominent examples of this category include Kuramoto phase oscillator \cite{acebron2005kuramoto} and Hopf oscillator \cite{righetti2008pattern}.

Since one of the motivations behind this work is to utilize the simplicity of CPG model to improve the training process of robot locomotion, the Hopf oscillator is chosen as the signal generator. The CPG model is composed of six coupled Hopf oscillators, each of which corresponds to one of the legs of the hexapod robot. The mathematical model of the Hopf oscillator is formulated as:
\begin{equation}
\left\{
\begin{aligned}
\dot{x}&=\alpha(\mu^2-x^2-y^2)x-\omega y \\
\dot{y}&=\beta(\mu^2-x^2-y^2)y+\omega x 
\end{aligned}
\right.
\label{hopf}
\end{equation}
where $x$ and $y$ are two state variables of oscillator, $\mu$ and $\omega$ are the amplitude and the frequency of the generated signals respectively, $\alpha$ and $\beta$ determine the rate that oscillator converges to the limit circle. In this paper, both $\alpha$ and $\beta$ are chosen to be 100.

The synchronized movements of six legs are realized though phase coupling between oscillators. Similar to the CPG model proposed in \cite{ouyang2021adaptive}, the formulation of coupled model used in this paper is given as follows:
\begin{equation}
\left\{
\begin{aligned}
\dot{x_i}&=\alpha(\mu^2-x_i^2-y_i^2)x_i-\omega y_i-k\sum_j \Delta_{ij,x} \\
\dot{y_i}&=\beta(\mu^2-x_i^2-y_i^2)y_i+\omega x_i+k\sum_j \Delta_{ij,y}
\end{aligned}
\right.
\label{coupled_hopf}
\end{equation}

In \eqref{coupled_hopf}, $i=1,...,6$, $k$ is the coefficient that adjusts coupling strength and $\Delta_{ij}$ refers to the coupling term, which is defined as:
\begin{figure*}[htbp]
\centerline{\includegraphics[width=0.75\textwidth]{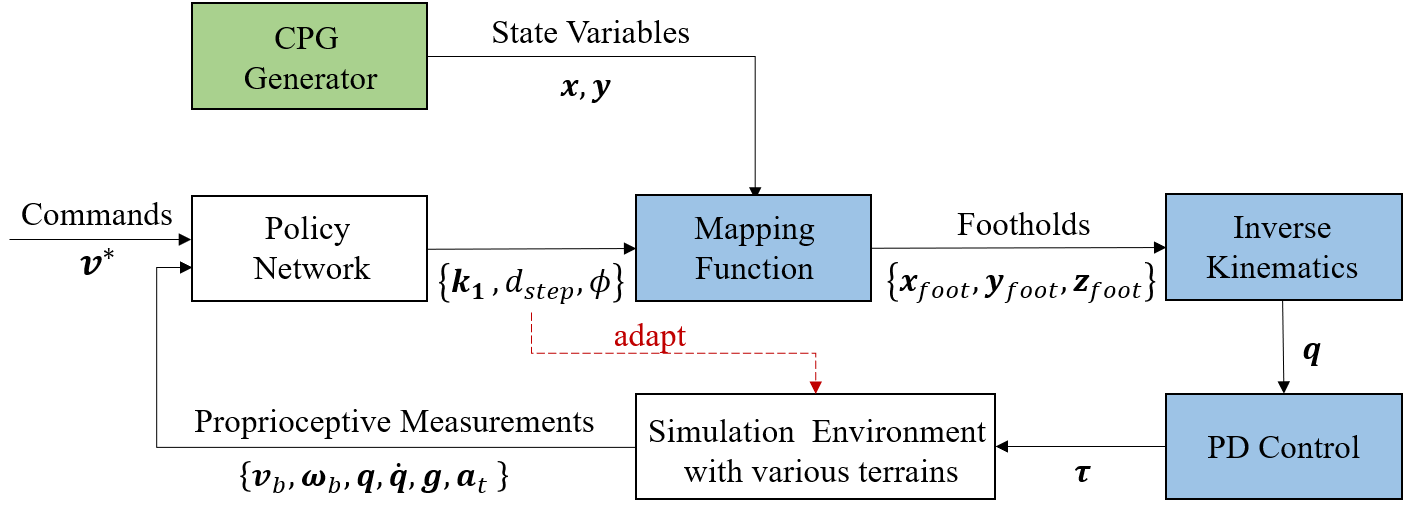}}
\caption{Schematic diagram of the proposed terrain-adaptive locomotion control method. Given proprioceptive observation from Issac Gym, the policy network output determines the parameters of the mapping function, which is used to transform CPG state variables $\mathbf{x}$ and $\mathbf{y}$ into foot positions. Following this, desired joint positions $\mathbf{q}$ are obtained using inverse kinematics, and the PD controller is used to produce torques $\mathbf{\tau}$ that actuate the robot.} 
\label{framework}
\end{figure*}
\begin{equation}
\left\{
\begin{aligned}
\Delta_{ij,x}&=-\sin\theta_{ij}\frac{x_j+y_j}{\sqrt{x_j^2+y_j^2}} \\
\Delta_{ij,y}&=\cos\theta_{ij}\frac{x_j+y_j}{\sqrt{x_j^2+y_j^2}}
\end{aligned}
\right.
\label{couling_term}
\end{equation}
where $\theta_{ij}$ is the phase difference between different oscillators. It also represents the phase difference between robot legs. For a hexapod robot, its typical gaits include tripod gait and quadruped gait. Only tripod locomotion is considered in this work, its corresponding phase differences is illustrated in Fig.\ref{leg_phase}. Specifically, the phase of the left front leg (LF) is utilized as the reference point, with its phase $\varphi_{LF}$ set to $0$.

\subsection{Reinforcement Learning}
The process of a hexapod robot learning locomotion skills can be modeled as a Markov decision process (MDP), which by definition can be expressed as a tuple $\{\mathcal{S},\mathcal{A},\mathcal{P},\mathcal{R}\}$. At time $t$, the robot obtains observation $o_t$ from the environment state $s_t \in \mathcal{S}$, interacts with the environment through action $a_t \in \mathcal{A}$ and receives reward $r_t=\mathcal{R}(s_t,a_t,s_{t+1}):\mathcal{S}\times\mathcal{A}\times\mathcal{S}\rightarrow\mathbb{R}$. The environment responds to this action and moves on to the next state $s_{t+1}$. The transition function $\mathcal{P}(s_{t+1}|s_t,a_t)$ depicts the possibility of transiting from state $s_t$ to next state $s_{t+1}$ after taking action $a_t$.  After a series of interactions between agent and environment, the expected return is defined as the sum of discounted rewards $R_t=\sum_{i=t} \gamma ^{i-t} r(s_i,a_i)$, where $\gamma \in (0,1)$ is the discount rate. The goal of reinforcement learning is to learn a policy that maximizes the cumulative reward.

\section{Adaptive learning of central pattern generator}
\label{details}
This section explains the realization of adaptive learning of central pattern generator. Fig. \ref{framework} presents the overall framework of the terrain-adaptive locomotion control method, which comprises three essential components, including the CPG generator, reinforcement learning module and assistive mapping functions that connect the former parts. The details will be explained below.
\subsection{Action Space}
In Section \ref{Background}, \eqref{coupled_hopf} gives the coupling state variables of the Hopf oscillator. Before implementing actual motion control, a mapping function is needed to transform these state variables into desired foot end positions in hip frame. Subsequently, inverse kinematics can be utilized to calculate the desired joint positions. The mapping function is defined as:
\begin{equation}
\left\{
\begin{aligned}
x_{i,foot}&=-d_{step}x_i\cos\phi \\
y_{i,foot}&=-d_{step}x_i\sin\phi \\
z_{i,foot}&=
\left\{
\begin{aligned}
-h&+k_1y_i \quad if y_i >0 \\
-h&+k_2y_i \quad otherwise 
\end{aligned}
\right.
\end{aligned}
\right.
\label{mapping_func}
\end{equation}
where $d_{step}$ is the step length, $h$ is the height of robot base, $k_1$ is the maximum distance between foot end and terrain surface that can be achieved though swing phase, $k_2$ is the maximum depth to which robot foot can sink into the ground during the stance phase, and $\phi$ controls the orientation of the robot body.

In the proposed reinforcement learning framework, action is defined as:
\begin{equation}
\{k_{1,i},d_{step},\phi\}, i=1,...,6
\label{action}
\end{equation}

This means in \eqref{mapping_func} only $h$ and $k_2$ have constant values, while the other parameters can be learnt and adjusted to adapt to external environment. Specifically, the height of the toe of each leg from the ground $k_{1,i}, i=1,...,6$ is output by the policy network, which greatly improves the ability of the robot to tackle uneven terrain. Besides, unlike common RL methods that directly map sensory information to joint positions (18-dimensional for a hexapod robot), learning the mapping function parameters reduces the dimensionality of the action space, making the search process easier during training.

\subsection{Observation Space}
Only proprioceptive measurements are considered in this work, including body information (base linear velocity $\mathbf{v}_b$ and base angular velocity $\mathbf{\omega}_b$), joint information (joint position $\mathbf{q}_i$ and joint velocity $\dot{\mathbf{q}}_i$ ), projected gravity vector $\mathbf{g}$, last actions predicted by the policy network $\mathbf{a}_t$ and velocity commands to track. The velocity commands include linear velocity $v^*_{b,x}$, $v^*_{b,y}$ in $x$, $y$ direction and the desired yaw rate in the body frame $w^*_{b,z}$. Therefore, the observation $\mathbf{o}_t$ is formulated as:
\begin{equation}
\{\{\mathbf{v}_b,\mathbf{\omega}_b\},\{\mathbf{q}_i,\dot{\mathbf{q}}_i\},\mathbf{g},\mathbf{a}_t,\{v^*_{b,x},v^*_{b,y},w^*_{b,z}\}\}, i=1,...,6
\label{observation}
\end{equation}

\subsection{Reward}
In this paper, reward is designed to track the linear velocity commands $v^*_{b,x}$, $v^*_{b,y}$ and yaw rate $w^*_{b,z}$ while keeping the natural and stable movement of the hexapod robot with minimal energy consumption. The reward function is expressed as a weighted sum of multiple reward terms. Table \ref{reward} shows the mathematical expressions of these reward terms and their corresponding weights, where $\phi(x)=e^{-\frac{||x||^2}{0.25}}$.

The above reward design is shared by plain DRL experimental settings and the proposed method that combines CPG and DRL. Also, experiments are performed to demonstrate that the introduction of CPG can help relieve the burden of reward design. More details will be given in Section \ref{Experiments}.

\section{Experiments}
\label{Experiments}
In this section, training details and experimental results of the proposed terrain-adaptive method are reported. Several key points to be validated regarding the proposed method include the ability to accomplish motion generation of hexapod robots, better terrain adaptability compared with plain CPG and superiority over end-to-end DRL in convergence rate and reward design complexity.
\begin{table}
\caption{Expressions of Reward Terms}
\begin{center}
\resizebox{8.5cm}{2.2cm}{
\begin{tabular}{|c|c|c|c|}
\hline
\cline{2-4}
\textbf{Term} & \textbf{Expression}& \textbf{Reward 1}& \textbf{Reward 2}\\
\hline
Linear velocity tracking & $\phi(v^*_{b,xy}-v_{b,xy})$ & 4$dt$ & 4$dt$\\
\hline
Angular velocity tracking & $\phi(\omega^*_{b,z}-\omega_{b,z})$ & 1$dt$ & 1$dt$\\
\hline
Linear velocity penalty & $||v_{b,z}||^2$ & -1$dt$ & $\times$\\
\hline
Angular velocity penalty & $||\omega_{b,xy}||^2$ & -0.05$dt$ & $\times$\\
\hline
Hip rotation penalty & $||\mathbf{q}_{hip}||^2$ & -0.5$dt$ & $\times$\\
\hline
Joint velocity penalty & $||\mathbf{\dot{q}}_{hip}||^2$ & -0.001$dt$ & $\times$\\
\hline
Joint acceleration penalty & $||\mathbf{\Ddot{q}}_{hip}||^2$ & -2.5e-7$dt$ & $\times$\\
\hline
Action rate penalty & $||\mathbf{a}_t-\mathbf{a}_{t-1}||^2$ & -0.01$dt$ & $\times$\\
\hline
Torque penalty & $||\mathbf{\tau}||^2$ & -1e-4$dt$ & -1e-4$dt$\\
\hline
Collision penalty & $F_{hip,leg,base}$ & -1$dt$ & -1$dt$\\
\hline
Feet air time & $\sum(t_{air}-0.5)$ & 1$dt$ & $\times$\\
\hline
\end{tabular}
}
\label{reward}
\end{center}
\end{table}
\begin{table}
\caption{PPO Hyperparameters}
\begin{center}
\begin{tabular}{|c|c|}
\hline
\cline{2-2}
\textbf{Parameter} & \textbf{Value} \\ \hline
Batch size & 98304 (4096x24)\\ \hline
Mini-batch size & 24576 (4096x6) \\ \hline
Number of epochs & 5\\ \hline
Clip range & 0.2\\ \hline
Entropy coefficient & 0.01\\ \hline
Discount factor & 0.99\\ \hline
GAE discount factor & 0.95\\ \hline
Desired KL-divergence & 0.01\\ \hline
Learning rate & adaptive\\ \hline
Number of hidden units & [512,256,128] \\ \hline
\end{tabular}
\label{hyperparameter}
\end{center}
\end{table}
\subsection{Training Details}
The proposed algorithm is implemented on a hexapod robot in NVIDIA Isaac Gym\cite{rudin2022learning} simulation environment. Isaac Gym allows for massive parallel training, where thousands of robots can be simulated and trained in parallel on a single GPU. In our case, the number of robots trained simultaneously is $4096$.

The deep reinforcement learning algorithm used in this work is Proximal Policy Optimization (PPO). The actor and critic network share the same architecture, with three hidden layers and ELU activation. Table \ref{hyperparameter} provides detailed information on the network and hyperparameters. Parameters of mapping function \eqref{mapping_func} are updated by the policy network at $200$ Hz, and the state variables $x$ and $y$ in oscillator function \eqref{coupled_hopf} are integrated at a frequency of $1000$ Hz. Consequently, the frequency of calculating torques from desired joint positions is $1000$ Hz as well.

\subsection{Experiment Results}
We conduct experiments in two settings: on flat ground and in a curriculum learning setting. In the curriculum learning structure, the global terrain map is created in a way similar to the one described in \cite{rudin2022learning}, where $20\times10$ local terrain blocks make up the global map, each with a different terrain type and difficulty level. The terrain types and range of difficulties include: 
\begin{itemize}
\item Random uniform terrain, where terrain heights are randomly sampled from $[-0.03,0.03]$ m.
\item Wave terrain, which is composed of sine waves with maximum amplitude gradually increasing from $0$ m to $0.45$ m.
\item Sloped terrain, of which slope varies from $0^{\circ}$ to $15^{\circ}$.
\end{itemize}

The robots are initialized on one side of the global map with random heading direction and velocity commands. The linear velocity command in $x$ direction ranges from $0.8$ m/s to $1$ m/s, and yaw rate varies in $[-0.1,0.1]$ m/s. To prevent unnatural lateral translation, the linear velocity of robots in the $y$ direction is restricted to be $0$. As the training proceeds, robots will learn to track the desired velocities and gradually spread across the entire map, eventually dispersing into terrain blocks with higher difficulty levels.

\begin{figure}
\centerline{\includegraphics[width=0.5\textwidth]{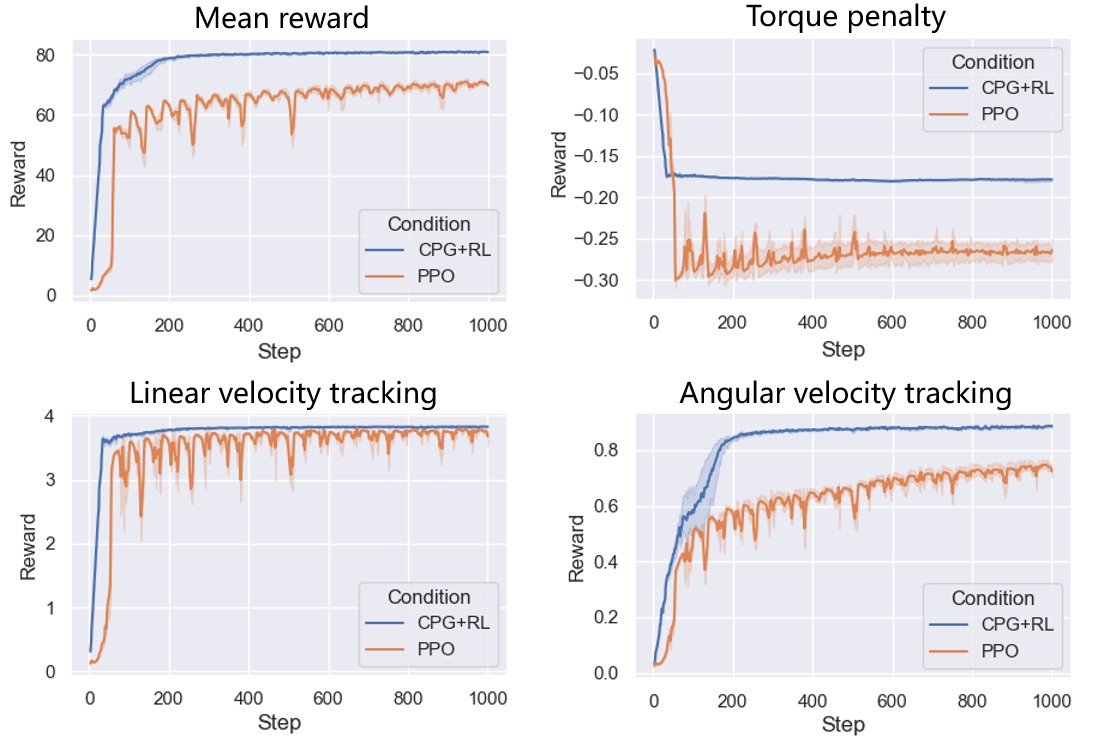}}
\caption{Average and standard deviation of the mean reward of CPG+RL and PPO on flat ground (over 4 runs, each with 1000 policy updates). The top right, bottom left and bottom right images are the learning curves of three dominant reward terms.}
\label{flat_reward}
\end{figure}

Motion verification:
To evaluate the impact of integrating CPG into RL framework, we compare the training results of the RL baseline and the proposed method on flat ground. Both methods utilize the same RL algorithm and reward terms. The RL baseline directly outputs the foot end positions in the hip frame of each leg, followed by inverse kinematics calculation and PD control. As illustrated in Fig. \ref{flat_reward}, the proposed method clearly outperforms the RL baseline in both convergence rate and final performance. The incorporation of CPG serves as prior knowledge that provides gait information for the robot and guarantees the periodicity of legged locomotion, accelerating the convergence of the training process. The improvement on the maximum average return is mainly due to three reward terms: torque penalty, linear velocity tracking and angular velocity tracking. As shown in the top right, bottom left and bottom right images of Fig. \ref{flat_reward}, the proposed algorithm is able to better track velocity commands with less energy consumption.

\begin{figure}
\centerline{\includegraphics[width=0.4\textwidth]{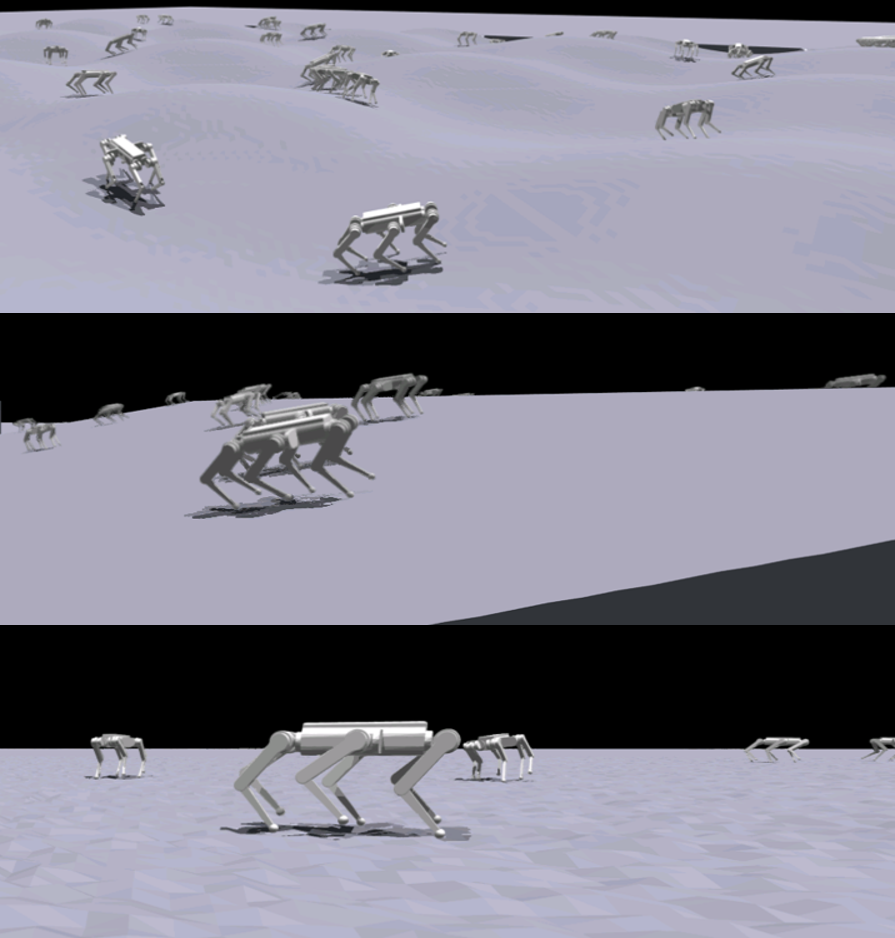}}
\caption{Robots walking on different terrains. The maximum height difference of the random uniform terrain is $0.06$ m. For wave terrain, the amplitude of each wave is $0.5$ m. For sloped terrain, the slope is 0.25 ($15^{\circ}$ in degree).}
\label{terrain}
\end{figure}
\begin{figure}
\centerline{\includegraphics[width=0.45\textwidth]{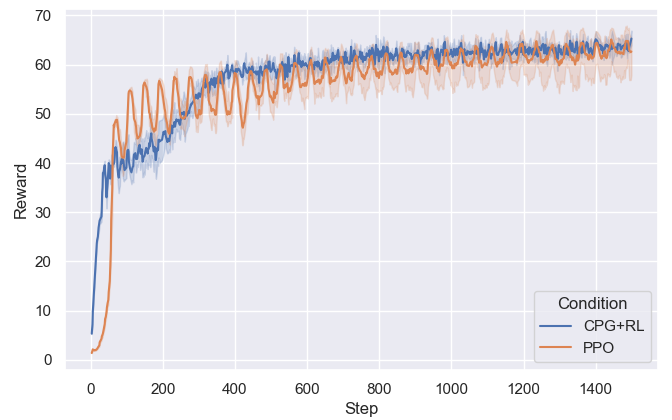}}
\caption{Average and standard deviation of the mean reward of CPG+RL and PPO on tough ground (over 4 runs, each with 1500 policy updates).}
\label{tough_reward}
\end{figure}
\begin{figure}
\centerline{\includegraphics[width=0.42\textwidth]{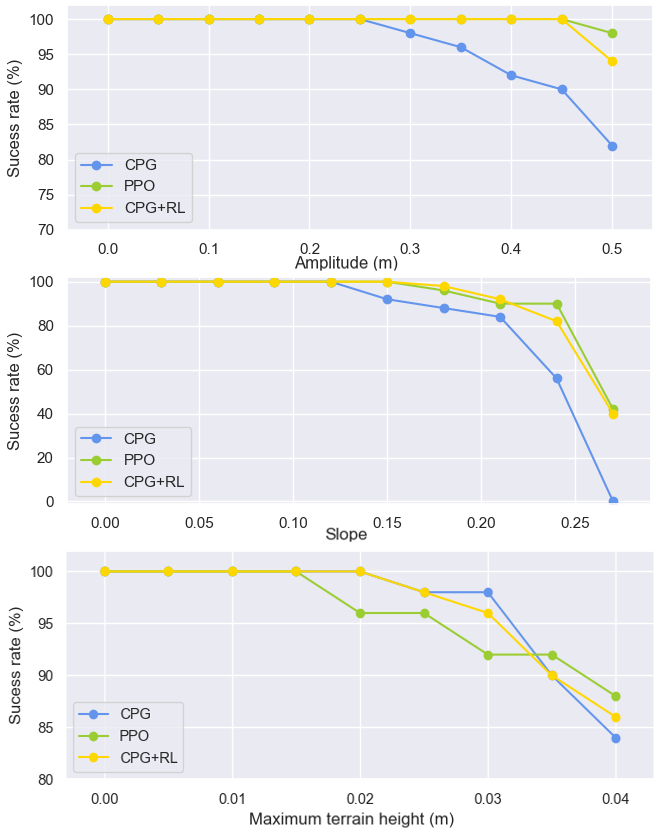}}
\caption{Success rate of three methods on terrains with increasing complexities. From top to bottom are the results on wave terrain, sloped terrain and random uniform terrain.}
\label{success_rate}
\end{figure}

Adaptation to uneven terrains: In order to test the terrain-adaptability of different terrains, further experiments are conducted on random uniform terrain, sloped terrain, and wave terrain. The snapshot of the hexapod robot traversing through these terrains under the guidance of the proposed method is displayed in Fig. \ref{terrain}. We compare the performance of the proposed method with a baseline RL method trained in a curriculum setting. As shown in Fig. \ref{tough_reward}, the learning curve of CPG+RL exhibits better robustness than that of RL baseline. Naturally, both methods shows a slightly degraded performance compared with the result on flat ground, with the maximum reward acquired being approximately 65. The success rate of tested algorithms on different terrains are displayed in Fig. \ref{success_rate}, where success is defined as no collision happens between the robot body and the ground during one episode of simulation (20 seconds). It is found that the CPG+RL method outperforms the CPG method on wave terrain and sloped terrain, which substantiates the enhanced terrain-adaptability of the proposed algorithm. However, no significant performance difference was observed between the three approaches on random uniform terrain.

\begin{figure}
\centerline{\includegraphics[width=0.5\textwidth]{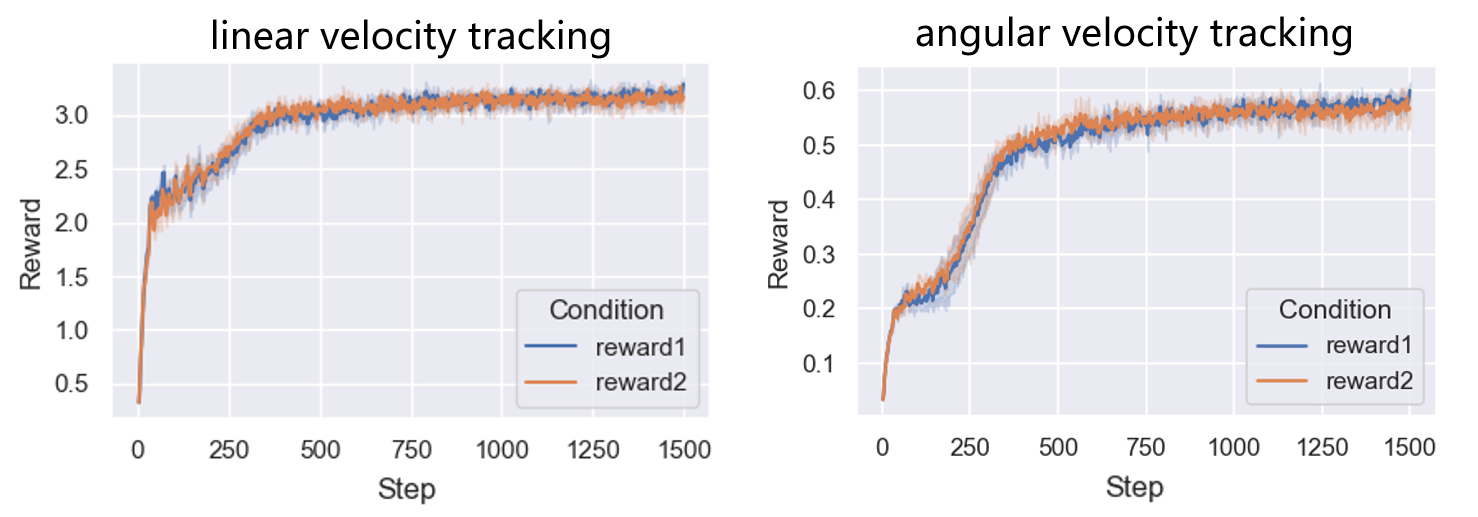}}
\caption{Results of using CPG+RL on rough terrains with different reward functions (averaged over 4 runs, each with 1500 policy updates).}
\label{reward_compare}
\end{figure}

Simple reward design: As discussed in Section \ref{introduction}, troublesome reward design is one of the most annoying deficiency when using reinforcement learning algorithms to learn end-to-end locomotion control. Since Fig. \ref{success_rate} has proved that the performance of CPG+RL is in line with that of RL in terrain-adaptability, we will further examine on the reliance of these two methods on reward design by comparing the results of tracking velocity commands using different reward terms. Table \ref{reward} shows two groups of reward terms, \textit{reward1} and \textit{reward2}. In \textit{reward2} setting, only \textit{velocity tracking (linear and angular)}, \textit{torque penalty} and \textit{collision penalty} term are considered. The experimental result, as illustrated in Fig. \ref{reward_compare}, shows that the two reward curves almost completely coincide, indicating that despite adopting a simpler reward design and reducing the constraints on robot locomotion, the capability of the proposed method to follow velocity commands remains unaffected. While in reinforcement learning framework, the simplification of reward terms will lead to unnatural and stumbling movement. These findings demonstrate the superiority of the proposed method over RL methods and further imply that it can reduce both the difficulty and time required for reward design while achieving the desired performance. 
\section{Conclusion}
\label{Conclusion}
This paper aims to introduce a terrain-adaptable method for hexapod locomotion control. Inspired by central pattern generators in biological motor control, the proposed method exploits multiple Hopf oscillators to generate synchronized movements and incorporates reinforcement learning to enhance adaptability towards uneven terrains by modulating the foothold planning parameters. Experimental results demonstrate its superiority over CPG method on wave and sloped terrains. Additionally, the proposed approach is found to exhibit less reliance on reward design, more natural locomotion gait and faster convergence rate than common RL methods. Future research will consider combining CPG with high-level decision-making to further improve the performance of the proposed method on more challenging scenarios.

\bibliographystyle{IEEEtran}
\bibliography{reference}

\end{document}